\documentclass[wcp]{jmlr}

\usepackage{microtype}
\usepackage{graphicx}
\usepackage{booktabs} 

\usepackage{hyperref}

\usepackage[font=small,skip=4pt]{caption}

\usepackage{amssymb}
\usepackage{mathtools}
\usepackage{algorithm,algpseudocode}

\usepackage[capitalize,noabbrev]{cleveref}

\usepackage{xspace}
\usepackage{paralist}

\usepackage[nolist]{acronym}
\begin{acronym}[UML]
	\acro{AI}{Artificial Intelligence}
 	\acro{ASWA}{Adaptive Stochastic Weight Averaging}

	\acro{CL}{Concept Learning}
	\acro{CEL}{Class Expression Learning}
	\acro{CWA}{Close World Assumption}
	\acro{CSV}{Comma-Separated Values}
	\acro{CNN}{Convolutional Neural Network}
	\acro{DL}{Description Logic}
	\acro{DL}{Deep Learning}
	\acro{DNN}{Deep Neural Network}
    \acro{FOL}{First-Order Logic}
  	\acro{GD}{Gradient Descent}
    \acro{KB}{Knowledge Base}
    \acro{KG}{Knowledge Graph}
    \acro{KGE}{Knowledge Graph Embedding}
	\acro{SW}{Semantic Web}
	\acro{OWA}{Open World Assumption}
	\acro{SGD}{Stochastic Gradient Descent}
 	\acro{SWA}{Stochastic Weight Averaging}
  
	\acro{MAP}{Maximum a Posteriori Probability Estimation}
	\acro{MRR}{Mean Reciprocal Rank}
	\acro{ML}{Machine Learning}
    \acro{MLE}{Maximum Likelihood Estimation}
    \acro{MDP}{Maximum Likelihood Estimation}
	\acro{MTL}{Multi-task Learning}
	\acro{MDP}{Markov Decision Process}
	\acro{NN}{Neural Network}
    \acro{KG}{Knowledge Graph}
    \acro{KGE}{Knowledge Graph Embedding}
    \acro{FGE}{Fast Geometric Ensembling}
\end{acronym}  

\newcommand{\approach}{\textsc{ASWA}\xspace}

\newcommand{\kg}{\ensuremath{\mathcal{G}}\xspace}
\newcommand{\entities}{\ensuremath{\mathcal{E}}\xspace}
\newcommand{\relations}{\ensuremath{\mathcal{R}}\xspace}

\newcommand{\RealNumbers}{\ensuremath{\mathbb{R}}\xspace}
\newcommand{\ComplexNumbers}{\ensuremath{\mathbb{C}}\xspace}
\newcommand{\Quaternions}{\ensuremath{\mathbb{H}}\xspace}

\newcommand{\emb}[1]{\ensuremath{\mathbf{e}_{#1}}}

\newcommand{\real}{\mathrm{Re}}

\usepackage[textsize=tiny]{todonotes}

\usepackage{longtable}

\usepackage{booktabs}

\usepackage{lineno}

\makeatletter
\let\Ginclude@graphics\@org@Ginclude@graphics 
\makeatother

\jmlrvolume{}
\jmlryear{}
\jmlrworkshop{ACML}

\title[Adaptive Stochastic Weight Averaging]{Adaptive Stochastic Weight Averaging}

  \author{\Name{Caglar Demir} \Email{
caglar.demir@upb.de}\\
   \Name{Arnab Sharma} \Email{arnab.sharma@upb.de}\\
   \Name{Axel-Cyrille Ngonga Ngomo} \Email{
axel.ngonga@upb.de}\\
   \addr Data Science Research Group, Paderborn University}

\begin{document}

\maketitle

\begin{abstract}
Ensemble models often improve generalization performances in challenging tasks. 
Yet, traditional techniques based on prediction averaging incur three well-known disadvantages: the computational overhead of training multiple models, increased latency, and memory requirements at test time.
To address these issues, the Stochastic Weight Averaging (SWA) technique maintains a running average of model parameters from a specific epoch onward. 
Despite its potential benefits, maintaining a running average of parameters can hinder generalization, as an underlying running model begins to overfit. 
Conversely, an inadequately chosen starting point can render SWA more susceptible to underfitting compared to an underlying running model.
In this work, we propose Adaptive Stochastic Weight Averaging (ASWA) technique that updates a running average of model parameters, only when generalization performance is improved on the validation dataset. 
Hence, ASWA can be seen as a combination of SWA with the early stopping technique, where the former accepts all updates on a parameter ensemble model and the latter rejects
any update on an underlying running model.
We conducted extensive experiments ranging from image classification to multi-hop reasoning over knowledge graphs.
Our experiments over 11 benchmark datasets with 7 baseline models suggest that ASWA leads to a statistically better generalization across models and datasets~\footnote{\url{https://github.com/dice-group/aswa}}.
\end{abstract}

\section{Introduction}
Ensemble learning is one of the most effective techniques to improve the generalization of \ac{ML} algorithms across challenging tasks~\citep{dietterich2000ensemble,murphy2012machine,goodfellow2016deep}.
In its simplest form, an ensemble model is constructed from a set of $K$ different learners over the same training set~\citep{breiman1996bagging}. 
At test time, a new data point is classified by taking a (weighted) average of $K$
the predictions of the $K$ learners~\citep{sagi2018ensemble}.
Although prediction averaging often improves the predictive accuracy, uncertainty estimation, and out-of-distribution robustness~\cite{garipov2018loss,liu2022deep},
it incurs three disadvantages: the computational overhead of training $K$ models and increased latency and memory requirements at test time~\citep{liu2022deep}.

Recently,~\citet{izmailov2018averaging} show that
\ac{SWA} technique often improves the generalization performance of a neural network by averaging $K$ \emph{snapshots} of parameters at the end of each epoch from a specific epoch onward.
Therefore, while a neural network being trained (called an underlying running model), an ensemble model is constructed through averaging the trajectory of an underlying running model in a parameter space.
Therefore, at test time, the time and memory requirements of using \ac{SWA} parameter ensemble is identical to the requirements of using a single neural network, e.g.,
e.g., the 160th of 200 training epoch across models on the CIFAR datasets.

Although the idea of averaging parameters of linear models to accelerate \ac{SGD} on convex problems dates back to~\citet{polyak1992acceleration}, 
~\citet{izmailov2018averaging} show that 
an effective parameter ensemble of a neural network architecture can be built by solely maintaining a running average of parameters from a specific epoch onward.
Yet, Selecting an unfittingly low number of training epochs leads \ac{SWA} to suffer more from underfitting compared to an underlying running model.
In other words, a parameter ensemble model is heavily influenced by the early stages of training.
Conversely, accepting all updates on a parameter ensemble model can hinder generalization, as an underlying running model begins to overfit.
Therefore, striking the right balance in the choice of training epochs is crucial for harnessing the full potential of \ac{SWA} in enhancing neural network generalization.

Here, we propose \ac{ASWA} technique that extends SWA by building a parameter ensemble based on an adaptive schema governed by the generalization trajectory on the validation dataset.
More specifically, while a neural network is being trained, \approach builds a parameter ensemble model by applying {\em soft}, {\em hard}, and {\em reject} updates:
\begin{compactitem}
    \item {\em Soft updates} refer to updating a parameter ensemble w.r.t. the parameters of running neural model to maintain a running average.
    \item {\em Hard updates} refer to reinitialization of a parameter ensemble model with parameters of a running model.
    \item {\em Reject updates} refer to the rejection of soft and hard updates.
\end{compactitem}
\approach can be seen as a combination of \ac{SWA} with the early stopping technique,
as \ac{SWA} accepts all soft updates on a parameter ensemble, 
while the early stopping technique rejects any updates on a running model. 
Although \approach introduces additional 
computations caused by monitoring the validation performance (as the early stopping technique does), \approach does not require any hyperparameter to be optimized on the validation dataset.
Our extensive experiments on 11 benchmark datasets for image classification, link prediction and multi-hop query answering with 7 state-of-the-art models suggest that 
\approach improves the generalization performance over \ac{SWA}, prediction averaging over multiple predictions, and early stopping across datasets and models. 
The main contributions of this paper are as follows:
\begin{compactenum}
    \item We propose a novel approach called \approach which improves over the existing parameter ensemble approach \ac{SWA} by exploiting the early stopping technique.
     \item We perform extensive evaluations considering different domains to quantify the effectiveness of \approach with using. 
     \item We provide an open-source implementation of \approach, including training and evaluation scripts along the experiment log files.
 \end{compactenum}

\section{Related Work \& Background}
\label{section:related_work}
\subsection{Ensemble Learning}
Ensemble learning has been extensively studied in the literature~\citep{bishop2006pattern,murphy2012machine}. 
In its simplest form, an ensemble model is constructed from a set of $K$ learners by averaging $K$ predictions~\citep{breiman1996bagging}.
At test time, a final prediction is obtained by averaging $K$ predictions of the $K$ learners.

For instance, averaging predictions of $K$ independently-trained neural networks of the same architecture can significantly boost the prediction accuracy over the test set. 
Although this technique introduces the computational overhead of training multiple models and/or increases latency and memory requirements at test time, it often improves the generalization performance in different learning problems~\citep{murphy2012machine}. 


Attempts to alleviate the computational overhead of training multiple models have been extensively studied.
For instance,~\citet{xie2013horizontal} show that saving parameters of a neural network periodically during training and composing a final prediction via a voting schema improves the generalization performance.
Moreover, the dropout technique can also be seen as a form of ensemble learning~\citep{hinton2012improving,srivastava2014dropout}. 
More specifically, preventing the co-adaptation of parameters by stochastically forcing them to be zero can be seen as
a \emph{geometric averaging}~\citep{warde2013empirical,baldi2013understanding,goodfellow2016deep}. 
Similarly, Monte Carlo Dropout can be seen as a variant of the Dropout that is used to approximate model uncertainty without sacrificing either computational complexity or test accuracy \citep{gal2016dropout}. 
\citet{garipov2018loss} show that
an optima of neural networks are connected by simple pathways having near constant training accuracy~\citet{draxler2018essentially}.
\citet{garipov2018loss} proposed FGE model to 
construct a ensemble model by averaging predictions of such neural networks is a promising means to improve the generalization performance.
Although the computational overhead of training multiple models can be alleviated by the aforementioned approaches, the increased latency and memory requirement remain unchanged at test time.
\subsection{Stochastic Weight Averaging}
~\citet{izmailov2018averaging} propose Stochastic Weight Averaging (SWA) that builds a high-performing parameter ensemble model having \emph{almost} the same training and exact test time cost of a single model. 
An SWA ensemble parameter update is performed as follows 
\begin{equation}
  \Theta_{\text{SWA}} \leftarrow \frac{\Theta_{\text{SWA}} \cdot n_{\text{models}} + \Theta}{n_{\text{models}} + 1} \,    
\end{equation}
where
$\Theta_{\text{SWA}}$ and $\Theta$ denote the parameters of the ensemble model and the parameters of the running model, respectively.
At each \ac{SWA} update, $n_{\text{models}}$ is incremented by 1.
Expectedly, finding a \emph{good} start epoch for SWA is important. Selecting an unfittingly low start point and total number of epochs can lead SWA to underfit. 
Similarly, starting \ac{SWA} only on the last few epochs does not lead to an improvement as shown in \cite{izmailov2018averaging}.
%
%
%
%
\subsection{Early Stopping}
The early stopping technique aims to avoid overfitting by stopping the training process before the convergence at training time.
It is considered as one of the simplest techniques to improve generalization performance~\citep{prechelt2002early}.
A safe strategy to find a good early stopping point is to monitor the performance on the validation dataset during training and consider the parameters leading to the lowest error rate on the validation set as the best parameter to be used at test time~\citep{muller2012regularization}.
Validation performances can be monitored at each epoch or according to an ad-hoc schema.
Expectedly, this yields a trade-off between a good generalization performance and the training runtime.

\subsection{Link Predictors \& Image Classifiers}

In our experiments, we evaluated the performance of \ac{SWA} with various neural network architectures in image classification and link prediction tasks.
In this subsection, we briefly elucidate neural network architectures. 
Most \ac{KGE} models are designed to learn continuous vector representations (\emph{embeddings}) of entities and relations tailored towards link prediction/single-hop reasoning~\citep{dettmers2018convolutional}.
They are often formalized as parameterized scoring functions $\phi_\Theta: \mathcal E \times \mathcal R \times \mathcal E \mapsto \mathbb R$,
where $\mathcal E$ denotes a set of entities, and $\mathcal R$ stands for a set of relations.
$\Theta$ often
consists of a d-dimensional entity embedding matrix
$\textbf{E} \in \mathbb{R}^{|\mathcal E| \times d}$
and a d-dimensional relation embedding matrix $\textbf{R} \in \mathbb R^{|\mathcal{R}| \times d}$. 
\Cref{table:models} provides an overview of selected \ac{KGE} models. 
Here, we focus on \ac{KGE} models based on multiplicative interactions, as recent works suggests that these multiplicative \ac{KGE} models often yield state-of-the-art performance if they are optimized well~\citep{ruffinelli2020you}. 
Moreover, these models perform well on multi-hop reasoning tasks~\citep{ren2023neural}. 

\begin{table}
\centering
\caption{Overview of KGE models. 
$\emb{}$ denotes an embedding vector, 
$d$ is the embedding vector size, 
$\overline{\emb{}} \in \ComplexNumbers$ corresponds to the complex conjugate of $\emb{.}$. 
$\times_n$ denotes the tensor product along the $n$-th mode.
$\otimes, \circ , \cdot$ stand for the Hamilton, Hadamard, and inner product, respectively.}
\label{table:models}
\resizebox{\columnwidth}{!}{\begin{tabular}{lcccccc}
  \toprule
  Model & Scoring Function & Vector Space & Additional\\
  \midrule
RESCAL~\cite{nickel2011three}&$\emb{h}\cdot\mathcal{W}_r \cdot\emb{t}$ &$\emb{h},\emb{t} \in \RealNumbers^d$  & $\mathcal{W}_r \in \RealNumbers^{d^2}$ \\
DistMult~\citep{yang2015embedding}      & $\emb{h} \circ \emb{r} \cdot \emb{t} $&$\emb{h},\emb{r},\emb{t} \in \RealNumbers^d$    &-\\
ComplEx~\citep{trouillon2016complex}    & $\real(\langle \emb{h}, \emb{r}, \overline{ \emb{t}} \rangle)$& $\emb{h},\emb{r},\emb{t} \in \ComplexNumbers^d$ &-\\
TuckER~\cite{balavzevic2019tucker}     & $\mathcal{W} \times_1 \emb{h} \times_2 \emb{r} \times_3 \emb{t}$& $\emb{h}, \emb{r},\emb{t} \in \RealNumbers^d$ &$\mathcal{W} \in \RealNumbers^{d^3}$ \\
QMult~\citep{demir2021hyperconvolutional} & $\emb{h} \otimes \emb{r} \cdot \emb{t}$ &$\emb{h},\emb{r},\emb{t} \in \Quaternions^d$&-\\
Keci \citep{demir2023clifford} & $\emb{h} \circ \emb{r} \cdot \emb{t}$ &$\emb{h},\emb{r},\emb{t} \in Cl_{p,q} (\mathbb{R}^d)$&-\\
  \bottomrule
\end{tabular}}
\end{table}

Recent results show that constructing an ensemble of \ac{KGE} models through prediction averaging improves the link prediction performance across datasets and \ac{KGE} models~\citep{demir2021hyperconvolutional,xu2021multiple}.
Yet, as $|\mathcal{E}| + |\mathcal{R}|$ grows, leveraging the prediction average technique becomes computationally prohibitive as it increases the memory requirement $K$ times, where $K$ denotes the number of models forming ensembles. 
To the best of our knowledge, constructing a parameter ensemble in the context of knowledge graph embeddings has not yet been studied. 
Hence, we perform parameter ensemble of knowledge graph embeddings which, to the best of our knowledge, not yet been studied.

We evaluate the performance of \ac{ASWA} with well-known neural network image classifiers, including VGG~\citep{simonyan2014very} and ResNet~\citep{he2016deep}.
Briefly, VGG consists of convolutional layers followed by max-pooling layers and topped with fully connected layers. 
The key characteristic of VGG is its deep architecture, with configurations ranging from VGG11 (11 layers) to VGG19. 
ResNet addresses the vanishing gradient problem associated with training very deep neural networks by introducing skip connections or shortcut connections. 
These connections allow the network to learn residual mappings, making it easier to optimize deeper architectures. ResNet is characterized by its residual blocks, where the input to a block is added to the output of the block before applying the activation function. 
ResNet architectures come in various depths, with deeper models achieving better performance on image classification and other computer vision tasks. In this work, we consider the ResNet model with 164 layers.
\section{Adaptive Stochastic Weight Averaging}
\label{section:methodology}

Here, we introduce our approach 
\approach defined as
\begin{equation}
\label{eq:aswa}
    \Theta_{\approach} = \sum_{i=1} ^N \pmb{\alpha}_i \odot \Theta_i,
\end{equation}
where $\Theta_i \in \mathbb{R}^d$ stands for a parameter vector of a running model at the $i-$th iteration.
$\odot$ denotes the scalar vector multiplication.
An ensemble coefficient $\pmb{\alpha}_i \in [0, 1]$ is a scalar value denoting the weight of $\Theta_i$ in $\Theta_\approach$.
An iteration can be chosen as an iteration over a training dataset
$\mathcal{D}_{\text{train}}$ over a mini-batch $\mathcal{B} \subset \mathcal{D}_{\text{train}}$, or a single data point $(\mathbf{x},y) \in \mathcal{D}_{\text{train}}$ at an epoch interval.
Here, $N$ denotes the number of epochs as does in \ac{SWA}~\citep{izmailov2018averaging}. 
Relationships between \approach and an underlying running model, the early stopping technique, and \ac{SWA} can be explicitly shown:
\begin{enumerate}    
    \item $\Theta_\approach $ corresponds to a parameter vector learned by a selected optimizer (e.g. \ac{SGD} or Adam), if $\pmb{\alpha}_{0:N-1}= \mathbf{0}$ and $\pmb{\alpha}_{N}=1$. 
     
    \item $\Theta_\approach$  corresponds to a parameter vector learned  by an early stopping technique that terminates the training process at the $j$-th iteration, 
    $\pmb{\alpha}_{0:j-1} = \mathbf{0}$, $\pmb{\alpha}_{j}=1$ and $\pmb{\alpha}_{j+1:N}= \mathbf{0}$.
    
    \item $\Theta_\approach$ becomes equivalent of $\Theta_{\text{SWA}}$ starting from the first iteration, if $\pmb{\alpha} = \mathbf{\frac{1}{N}}$. 

    \item $\Theta_\approach$ becomes equivalent of $\Theta_{\text{SWA}}$ starting from the $j$-th iteration if $\pmb{\alpha}_{0:j}= \mathbf{0}$ and $\pmb{\alpha}_{j+1:N}= \frac{\mathbf{1}}{\mathbf{N-j}}$.
\end{enumerate}
Here, we elucidate \approach and the impact of the start point $j$. 
Let $\Theta_{0}$ denote randomly initialized $d$ dimensional parameter vector.
Then $\Theta_{1}$ can be defined as the difference between 
$\Theta_{0}$ and $T$ consecutive parameter updates on $\Theta_{0}$
\begin{equation}
     \Theta_{1}= \Theta_{0} - \sum_{t=1} ^T \eta_{(0,t)} \nabla_{\Theta} \mathcal{L}_{\mathcal{B}_{(0,t)}} \big(\Theta_{(0,t)}\big),
     \label{eq:sgd_update_per_epoch}
 \end{equation}
where $T$ denotes the number of mini-batches to iterate over the training dataset $\mathcal{D}_{\text{train}}$. 
$\nabla_\Theta \mathcal{L}_{\mathcal{B}_{(0,t)}}$ denotes the gradients of the loss on the bases of the random mini-batch w.r.t. $\Theta_{(0,t)}$ at the $t$-th parameter update. 
$\eta_{(0,t)}$ and $\mathcal{L}_{\mathcal{B}_{(0,t)}} (\Theta_{(0,t)})$ denote the learning rate and the incurred mini-batch loss for the $t$-th parameter update, respectively. 
Let $T=N=2$, $\eta$ and $\Theta_0$ be given.
Then, $\Theta_\approach$ after the first iteration can be obtained as
\begin{equation}
\Theta_{\approach}^1 = \pmb{\alpha}_1 \odot \Theta_1
\label{eq:aswa_first_update}
\end{equation}
where $\Theta_1 = \big( \Theta_0 -(\eta_{(0,1)} \nabla_\Theta \mathcal{L}_{\mathcal{B}_{(0,1)}} + \eta_{(0,2)} \nabla_{\Theta} \mathcal{L}_{\mathcal{B}_{(0,2)}}) \big)$. 
Therefore, 
$\Theta_\approach$ at $N=2$ can be obtained as
\begin{equation}
\Theta_{\approach}^2 = (\pmb{\alpha}_1 \odot \Theta_1) + (\pmb{\alpha}_2 \odot \Theta_2),
\label{eq:aswa_second_update}
\end{equation}
where 
\begin{equation} 
\begin{split}
\Theta_2 = \Theta_0- \big(\eta_{(0,1)} \nabla_{\Theta} \mathcal{L}_{\mathcal{B}_{(0,1)}}+ \eta_{(0,1)} \nabla_\Theta \mathcal{L}_{\mathcal{B}_{(0,2)}} \\
+ \eta_{(1,1)} \nabla_{\Theta} \mathcal{L}_{\mathcal{B}_{(1,1)}} + \eta_{(2,1)} \nabla_{\Theta} \mathcal{L}_{\mathcal{B}_{(1,2)}} \big)
\end{split}
\end{equation}
%
With $\pmb{\alpha}_1=\pmb{\alpha}_2=1/2$, we obtained the following parameter ensemble
\begin{align*}
    \Theta_{\approach}^2=\Theta_0 - \Big( \eta_{(0,1)} \nabla_{\Theta} \mathcal{L}_{\mathcal{B}_{(0,1)}} + \eta_{(0,2)} \nabla_{\Theta} \mathcal{L}_{\mathcal{B}_{(0,2)}} + \\\frac{\eta_{(2,2)} \nabla_{\Theta} \mathcal{L}_{\mathcal{B}_{(1,1)}}}{2}  +
    \frac{\eta_{(2,2)}\nabla_{\Theta} \mathcal{L}_{\mathcal{B}_{(2,2)}}}{2} \Big),
    \label{eq:reducing_lr}
\end{align*}
where the influence of the running model seen at $N=2$ is less than the running model seen at $N=1$.
More generally, using \approach with equal ensemble weights $\pmb{\alpha}_{0:j}= \mathbf{0}$ and $\pmb{\alpha}_{j+1:N}= \frac{\mathbf{1}}{\mathbf{N-j}}$ can be rewritten as 
\begin{equation}
    \Theta_\approach=\Theta_j - \Big( 
    \sum_{i=j+1} ^{N} \sum_t ^T \frac{\eta_{(i,t)} \nabla_{\Theta}\mathcal{L}_{\mathcal{B}_{(i,t)}}}{i}\Big).
    \label{eq:reducing_lr_in_N_epoch}
\end{equation}
Therefore, as $i \to N$, the influence of the parameter updates on $\Theta_\approach$ decreases.
Recall that \Cref{eq:reducing_lr_in_N_epoch} corresponds to applying \ac{SWA} on the $j$-th epoch onward. 
Therefore, finding a suitable $j$ epoch to start maintaining a parameter average is important to alleviate underfitting. 
We argue that updating a parameter ensemble model with the parameters of an underlying running model suffering from overfitting may hinder the generalization performance.
Constructing a parameter ensemble by averaging over multiple nonadjacent epoch intervals 
(e.g. two nonadjacent epoch intervals $i$ to $j$ and $k$ to m s.t. $m \ge k+1 \ge j+1 \ge i$) 
may be more advantageous than selecting a single start epoch. 
With these considerations, we argue that determining $\pmb{\alpha}$ according to an adaptive schema governed by the generalization trajectory on the validation dataset can be more advantageous than using a prefixed schema.

\subsection{Determining Adaptive Ensemble Weights $\pmb{\alpha}$}
\label{subsec:determining_weights}

Parameter ensemble weights $\pmb{\alpha}$ can be determined in a fashion akin to early stopping technique.
More specifically, the trajectory of the validation losses can be tracked at end of the each epoch~\citep{prechelt2002early}.
By this, as the generalization performances degenerates, the training can be stopped.
This heuristic is known as early stopping technique.

In~\Cref{alg:ASWA}, we describe \approach with soft ensemble updates and rejection according to the trajectories of the validation performance of a running model $\Theta$ and a parameter ensemble model $\Theta_\approach$. 
By incorporating the validation performance of the running model, hard ensemble updates can be performed, i.e.,
if the validation performance of a running model is greater than the validation performances of the current ensemble model and the look-head. 
Therefore, the hard ensemble update, restarts the process of maintaining the running average of parameters, whereas the soft update, updates the current parameter ensemble with the running average of parameters seen after the hard update.
The validation performance of $\Theta_\approach$ cannot be less than the validation performance of $\Theta$.
Hence, a possible underfitting  depending on $N$ is mitigated with the expense of computing the validation performances.
Importantly, the validation performance of $\Theta_\approach$ cannot be less than the validation performance of $\Theta_\text{SWA}$ due to the rejection criterion, i.e., a parameter ensemble model is only updated if the validation performance is increased.


\begin{algorithm}
\caption{Adaptive Stochastic Weight Averaging (ASWA)}
\label{alg:ASWA}
\KwIn{Initial model parameters $\Theta_0$, number of iterations $N$, training dataset $\mathcal{D}_{\text{train}}$, validation dataset $\mathcal{D}_{\text{val}}$}
\KwOut{$\Theta_{\text{ASWA}}$}
\SetKwComment{Comment}{/* }{ */}
\BlankLine

$\Theta_{\text{ASWA}} \leftarrow \Theta_0$ \tcp*{Initialize Parameter Ensemble}
$\pmb{\alpha} \leftarrow \mathbf{0.0}$ \tcp*{Initialize N coefficients}
$\text{val}_{\approach} \leftarrow -1$

\For{$i \leftarrow 0, 1, \ldots, N$}{
    $\Theta_{i+1} \leftarrow \Theta_i - \alpha \nabla \mathcal{L}(\Theta_i)$ \tcp*{Update Running Model}
    $\hat{\Theta}_\approach \leftarrow \frac{\Theta_\approach \odot (\sum_{j=i}^N \pmb{\alpha}_{j}) + \Theta_{i+1}}{(\sum_{j=i}^N \pmb{\alpha}_{j}) + 1}$ \tcp*{Look-Ahead}
    $\text{val}_{\hat{\Theta}_\approach} \leftarrow \text{Eval}(\mathcal{D}_{\text{val}}, \hat{\Theta}_\approach)$ 

    \If{$\text{val}_{\hat{\Theta}_\approach} > \text{val}_{\approach}$}{
        $\Theta_\approach \leftarrow \hat{\Theta}_\approach$ \;
        
        $\pmb{\alpha}_{i+1} \leftarrow 1.0$ \;
        
        $\text{val}_{\approach} \leftarrow \text{val}_{\hat{\Theta}_\approach}$ \;
    }
    \Else{
        Do not update $\Theta_{\text{ASWA}}$ \tcp*{Reject Update}
    }
}
\end{algorithm}

\subsection{Computational Complexity}
At test time, the time and memory requirements of \approach are identical to the requirements of conventional training as well as SWA.
Yet, during training, the time overhead of \approach is linear in the size of the validation dataset.
This stems from the fact that at each epoch the validation performances are computed.
Since \approach does not introduce any hyperparameter to be tuned, \approach can be more practical if the overall training and hyperparameter optimization phases are considered. 
\section{Experimental Setup}

\subsection{Datasets}
In our experiments, we used the standard benchmark datasets  for image classification, link prediction, and multi-hop query answering tasks.
For the image classification tasks, CIFAR 10 and CIFAR 100 datasets are used.
The CIFAR-10 dataset consists of 60000 32x32 color images in 10 classes, with 6000 images per class.  There are 50000 training images and 10000 test images.
Similar to CIFAR-10, the CIFAR-100 dataset has 100 classes.
For the link prediction and multi-hop query answering tasks, we used UMLS, KINSHIP, Countries S1, Countries S2, Countries S3, NELL-995 h25, NELL-995 h50, NELL-995 h100, FB15K-237, YAGO3-10 benchmark datasets
Overviews of the datasets and queries are provided in~\Cref{table:datasets} and 
\Cref{tab:query-types}, respectively.
\begin{table}[htb]
    \caption{An overview of datasets in terms of the number of entities $\mathcal{E}$, number of relations $\mathcal{R}$, and the number of triples in each split of the knowledge graph datasets.}
    \centering
    \small
    \setlength{\tabcolsep}{1.0pt}
    \label{table:datasets}
    \begin{tabular}{lccccc}
    \toprule
    \textbf{Dataset} & \multicolumn{1}{c}{$|\entities|$}&  $|\relations|$ & $|\kg^{\text{Train}}|$ & $|\kg^{\text{Validation}}|$ &  $|\kg^{\text{Test}}|$\\
    \midrule
    Countries-S1    &271    &2  &1111   &24 &24\\
    Countries-S2    &271    &2  &1063   &24 &24\\
    Countries-S3    &271    &2  &985   &24 &24\\
    UMLS           &135      &46  &5,216      &652     &661\\
    KINSHIP        &104      &25  &8,544      &1,068   &1,074\\
    NELL-995 h100  &22,411   &43  &50,314     &3,763   &3,746\\
    NELL-995 h75   &28,085   &57  &59,135     &4,441   &4,389\\
    NELL-995 h50   &34,667   &86  &72,767     &5,440   &5,393\\
    FB15K-237      &14,541   &237 &272,115    &17,535  &20,466 \\
    YAGO3-10       &123,182  &37  &1,079,040  &5,000   &5,000 \\
    \bottomrule
    \end{tabular}

\end{table}
%
\subsection{Training and Optimization}
We followed standard experimental setups in our experiments.
For the link prediction and multi-hop query answering,
we followed the experimental setup used in~\citet{trouillon2016complex, ruffinelli2020you,arakelyan2021complex,demir2023clifford}. 
We trained DistMult, ComplEx, and QMult \ac{KGE} models with the following hyperparameter configuration: 
the number of epochs $N \in \{128, 256, 300\}$, Adam optimizer with $\eta=0.1$,
batch size $1024$, and an embedding vector size $d=128$.
Note that $d=128$ corresponds to 128 real-valued embedding vector size, hence 64 and 32 complex- and quaternion-valued embedding vector sizes respectively.
We ensure that all models have the same number of parameters while exploring various $d$.
Throughout our experiments, we used the KvsAll training strategy.
We applied the beam search combinatorial search to apply pre-trained aforementioned \ac{KGE} models to answer multi-hop queries.
More specifically, we compute query scores for entities via the beam search combinatorial optimization procedure, we keep the top $10$ most promising variable-to-entity substitutions.
We follow the exact experimental setup used in ~\citep{izmailov2018averaging} to perform experiments considering the CIFAR datasets. 
Hyperparameters on CIFAR100 and CIFAR10 are taken from~\citet{izmailov2018averaging}. 
We used the best found hyperparameters of SWA for \approach.
By this, we aim to observe whether \approach performs well regardless of carefully selecting the start point.
For the image classification experiments, we followed the experimental setup used in~\citet{izmailov2018averaging}. Herein, we compare our approach \ac{ASWA} to \ac{SWA}, \ac{SGD} and \ac{FGE} approaches. \ac{FGE} method, introduced by~\citet{garipov2018loss} leverages cyclical learning rates to aggregate models that are spatially close to each other, however, generating diverse predictions. This approach enables the creation of ensembles without added computational overhead. \ac{SWA} is inspired by FGE (as claimed by the authors), aiming to find a single model approximating an FGE ensemble while offering interpretability, convenience, and scalability. Therefore, we consider both of approaches in our evaluation.

We conducted the experiments three times and reported the mean and standard deviation performances.
Note that \approach is only trained on $95\%$ of the training dataset and the remaining $5\%$ is used as a proxy for the validation 
performance. 
%
%

\subsection{Evaluation}
To evaluate the link prediction and multi-hop query answering performances, we used standard metrics filtered mean reciprocal rank (MRR) and Hit@1, Hit@3, and Hit@10.
To evaluate the multi-hop query answering performances, we followed the complex query decomposition framework~\citep{arakelyan2021complex,demir2023litcqd}.
Therein, a complex multi-hop query is decomposed into subqueries, where the truth
value of each atom is computed by a pretrained knowledge graph embedding model/neural link predictor. 
Given a query, a prediction is obtained by
ranking candidates in descending order of their aggregated scores.
For each query type, we generate 500 queries to evaluate the performance.
For the image classification task, we applied the standard setup~\citep{krizhevsky2009learning} 
and repeated SWA experiments~\citep{izmailov2018averaging}.

\begin{table}[htb]
\caption{Overview of different query types. 
Query types are taken from \cite{demir2023litcqd}.}
\label{tab:query-types}
    \centering
    \setlength{\tabcolsep}{1.0pt}
    
    \begin{tabular}{@{}ll@{}}
        \toprule
        \multicolumn{2}{@{}c@{}}{\bfseries Multihop Queries} \\
        \midrule

        2p            & $E_?\:.\:\exists E_1:r_1(e,E_1)\land r_2(E_1, E_?)$                                \\
        3p            & $E_?\:.\:\exists E_1E_2.r_1(e,E_1)\land r_2(E_1, E_2)\land r_3(E_2,E_?)$           \\
        2i            & $E_?\:.\:r_1(e_1,E_?)\land r_2(e_2,E_?)$                                           \\
        3i            & $E_?\:.\:r_1(e_1,E_?)\land r_2(e_2,E_?)\land r_3(e_3,E_?)$                         \\
        ip            & $E_?\:.\:\exists E_1.r_1(e_1,E_1)\land r_2(e_2,E_1)\land r_3(E_1,E_?)$             \\
        pi            & $E_?\:.\:\exists E_1.r_1(e_1,E_1)\land r_2(E_1,E_?)\land r_3(e_2,E_?)$             \\
        2u            & $E_?\:.\:r_1(e_1,E_?)\lor r_2(e_2,E_?)$                                            \\
        up            & $E_?\:.\:\exists E_1.[r_1(e_1,E_1)\lor r_2(e_2,E_1)]\land r_3(E_1,E_?)$\\
        \bottomrule
    \end{tabular}
\end{table}

\subsection{Implementation}
We provide an open-source implementation of \approach, including training and evaluation scripts along the experiment log files~\footnote{\url{https://github.com/dice-group/aswa}}.
\section{Results}
\begin{table*}
\caption{Average accuracy ($\%$) of SWA, SGD, FGE, and \approach on the CIFAR datasets. 
\approach is trained only 95\% of the train data with best performing hyperparameters for SWA.}
\label{table:cifar10}
\centering
\small
\begin{tabular}{ccccccc}
    \toprule
	 & & & & & 	\\
& \textbf{\approach} & \textbf{SGD} & \textbf{FGE} & \textbf{SWA} \\
\hline
\multicolumn{6}{c}{\small CIFAR-100}\\
\hline
    VGG-16BN &$71.9 \pm 0.06$& 
    $72.55 \pm 0.10$ 
    & $74.26 \pm $ 
    & $\mathbf{74.27 \pm 0.25}$ \\
    ResNet-164 &$\mathbf{81.6 \pm 0.10}$&
    $78.49 \pm 0.36$ & $79.84\pm$ 
    & $80.35 \pm 0.16$ \\
    WRN-28-10  &$\mathbf{82.9\pm 0.20}$&
    $80.82 \pm 0.23$ & $82.27\pm$ & $82.15 \pm 0.27$ \\
    \hline 
    \multicolumn{6}{c}{\small CIFAR-10}\\
    \hline 
    VGG-16BN  &$\mathbf{94.63 \pm 0.20}$&
    $93.25 \pm 0.16$ 
    & $93.52\pm$
    & $93.64 \pm 0.18$ \\
    ResNet-164  &$\mathbf{96.27 \pm 0.01}$&
    $95.28 \pm 0.10$ & $95.45\pm$ 
    & $95.83 \pm 0.03$ \\
    WRN-28-10  &$96.64 \pm 0.10$&
    $96.18 \pm 0.11$ &$96.36\pm$
    &$\mathbf{96.79 \pm 0.05}$ \\
    \bottomrule   
\end{tabular}
\end{table*}

\begin{table*}[htb]
\caption{Average accuracy ($\%$) of SGD, SWA, \approach between 300 and 1000 epochs on CIFAR-10.
\approach is trained only 95\% of the train data with best performing hyperparameters for SWA reported.}
\label{table:cifar10_different_epochs}
\centering
\small
\resizebox{\textwidth}{!}{
\begin{tabular}{l ccccccccccccc ccccccccccc}
  \toprule
                & 300         & 400           & 500          & 600           & 700         &800           &900 & 1000&\\
   \toprule
ResNet164 & $93.7 \pm 0.4$ & $93.4 \pm 0.1$ & $93.4 \pm 0.3$ & $93.3 \pm 0.3$ & $93.5 \pm 0.2$ & $93.6 \pm 0.0$ & $93.9 \pm 0.3$ & $93.8 \pm 0.1$\\
ResNet164-SWA &$95.7 \pm 0.2$ & $95.9 \pm 0.2$ & $\mathbf{96.0 \pm 0.2}$ & $\mathbf{96.0 \pm 0.1}$ & $96.0 \pm 0.1$ & $96.0 \pm 0.1$ & $96.0 \pm 0.1$ & $96.0 \pm 0.1$\\
ResNet164-ASWA & $\mathbf{95.8 \pm 0.1}$ & $\mathbf{95.9 \pm 0.1}$ & $\mathbf{96.0 \pm 0.2}$ & $\mathbf{96.1 \pm 0.2}$ & $\mathbf{96.1 \pm 0.1}$ & $\mathbf{96.1 \pm 0.1}$ & $\mathbf{96.1 \pm 0.0}$ & $\mathbf{96.3 \pm 0.1}$\\
\midrule
VGG16 & $89.5 \pm 0.4$ & $89.4 \pm 0.9$ & $90.0 \pm 0.3$ & $90.2 \pm 0.3$ & $89.1 \pm 1.0$ & $90.1 \pm 0.2$ & $89.5 \pm 0.3$ & $89.8 \pm 0.3$ \\
VGG16-SWA & $92.5 \pm 0.2$ & $92.6 \pm 0.1$ & $92.6 \pm 0.2$ & $92.6 \pm 0.3$ & $92.4 \pm 0.2$ & $92.4 \pm 0.1$ & $92.4 \pm 0.2$ & $92.4 \pm 0.2$ \\
VGG16-ASWA & $\mathbf{92.7 \pm 0.2}$ & $\mathbf{92.7 \pm 0.2}$ &$\mathbf{92.8 \pm 0.2}$ & $\mathbf{92.8 \pm 0.1}$ & $\mathbf{92.7 \pm 0.2}$ & $\mathbf{92.7 \pm 0.2}$ & $\mathbf{92.8 \pm 0.2}$ & $\mathbf{92.8 \pm 0.2}$\\
\midrule
VGG16BN & $90.7 \pm 0.6$ & $90.5 \pm 0.3$ & $90.4 \pm 1.2$ & $90.5 \pm 0.6$ & $90.4 \pm 0.8$ & $90.8 \pm 0.3$ & $91.0 \pm 0.4$ & $90.8 \pm 0.3$\\
VGG16BN-SWA & $94.2 \pm 0.0$ & $94.3 \pm 0.1$ & $94.3 \pm 0.0$ & $94.4 \pm 0.1$ & $94.4 \pm 0.1$ & $94.4 \pm 0.0$ & $94.4 \pm 0.1$ & $94.4 \pm 0.1$\\
VGG16BN-ASWA & $\mathbf{94.3 \pm 0.1}$ & $\mathbf{94.5 \pm 0.1}$ & $\mathbf{94.5 \pm 0.1}$ & $\mathbf{94.6 \pm 0.0}$ & $\mathbf{94.6 \pm 0.0}$ & $\mathbf{94.7 \pm 0.1}$ & $\mathbf{94.7 \pm 0.1}$ & $\mathbf{94.6 \pm 0.2}$\\
\bottomrule
\end{tabular}}
\end{table*}
\begin{table*}
\caption{
Average accuracy of ASWA across different epochs on CIFAR-10 with train and val ratios.}
\centering
\small
\label{table:cifar10_aswa_ratios}
\resizebox{\textwidth}{!}{
\begin{tabular}{l ccccccccccccc ccccccccccc}
\toprule
            & 300           & 400         & 500          & 600           & 700          & 800         &900 & 1000&\\
\toprule
\multicolumn{10}{c}{90/10}\\
\toprule
VGG16 & $92.5 \pm 0.3$ & $92.6 \pm 0.2$ & $92.7 \pm 0.3$ & $92.7 \pm 0.2$ & $92.6 \pm 0.2$ & $92.6 \pm 0.2$ & $92.6 \pm 0.2$ & $92.7 \pm 0.2$\\
VGG16BN & $94.1 \pm 0.2$ & $94.3 \pm 0.2$ & $94.3 \pm 0.2$ & $94.4 \pm 0.2$ & $94.5 \pm 0.2$ & $94.5 \pm 0.2$ & $94.5 \pm 0.1$ & $94.5 \pm 0.1$\\
\toprule
\multicolumn{10}{c}{80/20}\\
\toprule
VGG16 & $92.1 \pm 0.1$ & $92.3 \pm 0.2$ & $92.3 \pm 0.2$ & $92.3 \pm 0.2$ & $92.3 \pm 0.1$ & $92.3 \pm 0.1$ & $92.3 \pm 0.0$ & $92.2 \pm 0.1$\\
VGG16BN & $93.8 \pm 0.1$ & $93.9 \pm 0.1$ & $93.8 \pm 0.1$ & $94.0 \pm 0.1$ & $94.1 \pm 0.1$ & $94.1 \pm 0.1$ & $94.1 \pm 0.1$ & $94.2 \pm 0.0$\\
\bottomrule   
\end{tabular}}
\end{table*}
\subsection{Image Classification Results}

Tables~\ref{table:cifar10},~\ref{table:cifar10_different_epochs}, and~\ref{table:cifar10_aswa_ratios} report the image classification performance on the CIFAR-100 and CIFAR-10 benchmark dataset.
By following the experimental setup of \cite{izmailov2018averaging}, we conduct the experiments three times and report the mean and standard deviation performances.
Important to know that \approach is only trained on $95\%$ of the training dataset and the remaining $5\%$ is used as a proxy for the validation performance.
As it can be seen in \Cref{table:cifar10}, although we used the best performing hyperparameters of SWA for \approach,
\approach on average outperforms SGD, FGE, and SWA, 1.3\%, 1.1\%, and 1\% accuracy respectively. \Cref{table:cifar10_different_epochs} shows that the generalization performance of a running model does not degenerate from 300 to 1000 epochs where \approach slightly improves the results over SGD and SWA.
For instance, the performance of ResNet164 does not change more than 0.1\%, whereas SWA and \approach slightly improve the results.

The benefits of \approach become less tangible over ASWA provided that hyperparameters of a running model are optimized against overfitting, e.g., learning rate scheduling and weight decay are optimized in~\cite{izmailov2018averaging}.
Due to the space constraint, we relegated other experiments to supplement materials. To quantify the impact of training and validation ratios on the performance of the \approach, we conducted two additional experiments.
\Cref{table:cifar10_aswa_ratios} show that VGG16BN with \approach trained only on the 80\% of the training dataset outperforms VGG16BN with SWA trained on the full training dataset.

\subsection{Link Prediction and Multi-hop Query Answering Results}
\Cref{table:wn18rr_fb15k237_yago310,table:lp_results_nell,table:query_answering_fb_results,table:train_fb15k237_yago310} report the link prediction and multi-hop query answering results on benchmark datasets.
Overall, experimental results suggest that SWA and \approach consistently lead to better generalization performance in all metrics than conventional training in the link prediction and multi-hop query answering tasks.
\Cref{table:wn18rr_fb15k237_yago310} shows that SWA and \approach outperform the conventional training on WN18RR, FB15k-237, and YAGO30 in 48 out of 48 scores.
Important to note that  \approach outperforms SWA in 37 out of 48 scores.
scores
\Cref{table:lp_results_nell} shows that \approach outperforms SWA in out 47 of 48 scores on NELL datasets.
%
\begin{table*}[htb]
\caption{Link prediction results on  WN18RR, FB15K-237 and YAGO3-10.}
\label{table:wn18rr_fb15k237_yago310}
\centering
\resizebox{\textwidth}{!}{
\begin{tabular}{l  cccc cccc cccc}
\toprule
  &\multicolumn{4}{c}{\textbf{WN18RR}} & \multicolumn{4}{c}{\textbf{FB15K-237}} & \multicolumn{4}{c}{\textbf{YAGO3-10}}\\
  \cmidrule(l){2-5} \cmidrule(l){6-9} \cmidrule(l){10-13}
                   & MRR  & @1     &@3     & @10 & MRR  & @1     &@3     & @10       & MRR  & @1 & @3 & @10\\
\toprule
DistMult          
&0.345&0.336&0.351&0.361  
&0.097&0.060&0.099&0.166
&0.148&0.102&0.165&0.235 \\
DistMult-SWA      
&\textbf{0.359}&\textbf{0.353}&\textbf{0.361}&\textbf{0.372}
&0.154&0.104&0.164&0.247
&0.343&0.265&0.383&0.497\\
DistMult-\approach
&0.354&0.349&0.356&0.363 
&\textbf{0.209}&\textbf{0.151}&\textbf{0.229}&\textbf{0.321} 
&\textbf{0.432}&\textbf{0.360}&\textbf{0.477}&\textbf{0.562}\\
\midrule
ComplEx           
&0.285 &0.258 &0.304&0.336 
&0.100 &0.063 &0.105&0.170 
&0.195&0.124&0.226&0.334   
\\
ComplEx-SWA      
&0.289&0.262&0.307&0.334 
&0.143&0.095&0.153&0.234 
&\textbf{0.391}&\textbf{0.310}&\textbf{0.441}&0.541 
\\
ComplEx-\approach 
&\textbf{0.354}&\textbf{0.343}&\textbf{0.360} 
&\textbf{0.315} & \textbf{0.206} & \textbf{0.148}&\textbf{0.223}&\textbf{0.315} 
&\textbf{0.391} &0.302&0.440&\textbf{0.548} 
\\
\midrule
QMult              
& 0.114 & 0.091 & 0.122 & 0.153 
& 0.105 &0.077&0.110&0.153       
&0.168  & 0.118 & 0.179 & 0.263 
\\
QMult-SWA         
&0.112 & 0.091 & 0.121 & 0.152 
&0.096 & 0.076 & 0.099 & 0.122 
&\textbf{0.398}&\textbf{0.329} &\textbf{0.439}&\textbf{0.525}     
\\
QMult-\approach   
& \textbf{0.152} &\textbf{0.121} &\textbf{0.164} &\textbf{0.214}    
& \textbf{0.251} & \textbf{0.184} & \textbf{0.273} & \textbf{0.380} 
&0.377 &0.310&0.417&0.507 
\\
\midrule
Keci              
&0.336&0.316&0.351&0.366    
&0.148&0.096&0.154&0.252    
&0.294&0.219&0.328&0.440    
\\
Keci-SWA          
&0.337&0.317&0.351&0.366    
&0.217&0.148&0.235&0.354    
&0.388&0.314 & 0.428 & 0.527 
\\
Keci-\approach    
&\textbf{0.355} &\textbf{0.348}&\textbf{0.358} & \textbf{0.368} 
& \textbf{0.256} &\textbf{0.184} &\textbf{0.279} &\textbf{0.398} 
& \textbf{0.397}&\textbf{0.322}&\textbf{0.441}&\textbf{0.543}
\\
\bottomrule
\end{tabular}}
\end{table*}

\begin{table*}[htb]
\caption{Link prediction results on the NELL-995 benchmark datasets.}
\label{table:lp_results_nell}
\centering
\resizebox{\textwidth}{!}{
\begin{tabular}{l  cccc cccc cccc}
\toprule
  &\multicolumn{4}{c}{\textbf{h100}} & \multicolumn{4}{c}{\textbf{h75}} & \multicolumn{4}{c}{\textbf{h50}}\\
  \cmidrule(l){2-5} \cmidrule(l){6-9} \cmidrule(l){10-12}
                   & MRR  & @1     &@3     & @10 & MRR  & @1     &@3     & @10       & MRR  & @1 & @3 & @10\\
\toprule
DistMult           & 0.152 & 0.112 & 0.166 & 0.228 & 0.151 & 0.108 & 0.164 & 0.236&0.133 &  0.094 &0.148 &0.205\\
DistMult-SWA       & 0.206 & 0.143 & 0.231 & 0.334 &0.168 & 0.127 &0.182&0.250& 0.198 & 0.148 & 0.220 & 0.290 \\
DistMult-\approach &\textbf{0.233} & \textbf{0.169} & \textbf{0.260} & \textbf{0.357}&\textbf{0.228}&\textbf{0.171}&\textbf{0.249}&\textbf{0.344}
& \textbf{0.245} & \textbf{0.180} & \textbf{0.273} & \textbf{0.371} \\
\midrule
ComplEx & 0.202    & 0.144 & 0.225 & 0.321          &0.200 & 0.142 & 0.223 & 0.312&0.208&0.150&0.231&0.323\\
ComplEx-SWA        & 0.202 & 0.143 & 0.221 & 0.319  &0.194 & 0.137 &0.214  &0.307 &0.232&\textbf{0.174}&0.256&0.346\\
ComplEx-\approach  &\textbf{0.228}&\textbf{0.164}&\textbf{0.250}&\textbf{0.359}&\textbf{0.223}&\textbf{0.160}&\textbf{0.246}&\textbf{0.346} &\textbf{0.236}&0.171&\textbf{0.263}&\textbf{0.361}\\
\midrule
QMult             & 0.155 & 0.100 & 0.173 & 0.261 & 0.168 & 0.115 & 0.184 & 0.275&0.174&0.119&0.193&0.283\\
QMult-SWA         &0.158 & 0.105 & 0.177 & 0.262
& 0.169 & 0.114 & 0.187 & 0.282
&0.191&\textbf{0.134}&0.212&0.305\\
QMult-\approach   
&\textbf{0.197}&\textbf{0.135}&\textbf{0.219}&\textbf{0.321} &\textbf{0.185}
&\textbf{0.126}&\textbf{0.206}&\textbf{0.296}
&\textbf{0.191} &0.130 &\textbf{0.213} &\textbf{0.309}\\
\midrule
Keci     &0.198&0.143&0.219&0.312    
         &0.188&0.131&0.206&0.303
         &0.203&0.153&0.222&0.301\\
Keci-SWA &0.208&0.153&0.231&0.316 
         &0.222&0.158&0.245&0.346&0.229&0.174&0.253&0.335\\
Keci-\approach    
& \textbf{0.257} & \textbf{0.182} & \textbf{0.287} & \textbf{0.406} &\textbf{0.231}&\textbf{0.165}&\textbf{0.255}&\textbf{0.361}
&\textbf{0.237}&\textbf{0.177}&\textbf{0.263}&\textbf{0.351}\\
\bottomrule
\end{tabular}}
\end{table*}

\Cref{table:query_answering_fb_results} reports the multi-hop query answering results on FB15k-237.
Results suggest that the benefits of using \approach become particularly tangible in answering 2 and up multi-hop queries involving $\land$ and $\lor$.
\begin{table*}[!htbp]
\centering
\small
\caption{Multi-hop query answering MRR results on FB15k-237.}
\label{table:query_answering_fb_results}
\begin{tabular}{lcccccccccc}
        \toprule
& \textbf{2p}&\textbf{3p}&\textbf{3i}       & \textbf{ip}       & \textbf{pi}       & \textbf{ 2u}       & \textbf{up}       \\
        \midrule
DistMult           &\textbf{0.007} &\textbf{0.007}&0.044 & 0.003 &0.097&0.020&0.006 \\
SWA       &0.003 &0.002 &0.106&0.004&\textbf{0.098}&0.031&\textbf{0.007} \\
\approach &0.002&0.003&\textbf{0.175}&\textbf{0.005}&0.092&\textbf{0.055}&0.002\\
\midrule
ComplEx            &\textbf{0.009}&0.001&0.036&0.003&0.092&0.014&\textbf{0.006}\\
SWA        &0.002&0.002&0.136&\textbf{0.005}&\textbf{0.105}&0.030&0.005\\
\approach  &0.002&\textbf{0.003}&\textbf{0.155}&\textbf{0.005}&0.100&\textbf{0.049}&0.001\\
\midrule
QMult              &0.002&0.003&0.034&0.000&0.099&0.014&0.006\\
SWA          &0.003&0.002&0.089&0.001&0.092&0.027&\textbf{0.007}\\
\approach    &\textbf{0.004}&\textbf{0.005}&\textbf{0.183}&\textbf{0.002}&\textbf{0.117}&\textbf{0.072}&0.003\\
\midrule
Keci              &\textbf{0.009}&0.005&0.052&0.002&0.085&0.027&0.013\\
SWA          &0.007         &0.005&0.101&0.002&0.097&0.059&\textbf{0.014}\\
\approach    &0.004         &0.005&\textbf{0.165}&\textbf{0.007}&\textbf{0.106}&\textbf{0.064}&0.004\\
\bottomrule
\end{tabular}
\end{table*}
\begin{table*}[t]
\caption{
Link prediction results on  the training splits of the two large link prediction datasets.}
\label{table:train_fb15k237_yago310}
\centering
\small
\begin{tabular}{l  cccc cccc cccc}
\toprule
  &\multicolumn{4}{c}{\textbf{FB15K-237}} & \multicolumn{4}{c}{\textbf{YAGO3-10}}\\
  \cmidrule(l){2-5} \cmidrule(l){6-9}
                   & MRR  & @1     &@3     & @10       & MRR  & @1 & @3 & @10\\
\toprule
DistMult           
&0.780&0.729&0.809&0.872 
&0.773&0.709&0.819&0.883
\\
DistMult-SWA       
&\textbf{0.802}&\textbf{0.749}&\textbf{0.834}&\textbf{0.901}
&\textbf{0.980}&\textbf{0.970}&\textbf{0.988}&\textbf{0.995}\\  
DistMult-\approach &0.497&0.399&0.548&0.684
&0.893&0.861&0.917&0.943\\
\midrule
ComplEx           
&0.771&0.725&0.794&0.858
&0.987&0.980&0.995&0.997
\\
ComplEx-SWA       
&\textbf{0.789}&\textbf{0.734}&\textbf{0.822}&\textbf{0.890}
&\textbf{0.998}&\textbf{0.998}&\textbf{1.000}&\textbf{1.000}
\\    
ComplEx-\approach &0.448&0.349&0.496&0.637
&0.925&0.899&0.945&0.967
\\
\midrule
QMult          
&\textbf{0.996}&\textbf{0.993}&\textbf{0.998}&0.999    
&1.000&1.000&1.000&1.000
\\
QMult-SWA      
&0.995&0.991&\textbf{0.998}&\textbf{1.000}
&1.000&1.000&1.000&1.000
\\
QMult-\approach  &0.666&0.568&0.725&0.843
& 0.977&0.963&0.993&0.997
\\
\midrule
Keci           
&0.758&0.702&0.791&0.863
&0.999&0.998&1.000&1.000\\
Keci-SWA       
&\textbf{0.822}&\textbf{0.755}&\textbf{0.870}&\textbf{0.945}
&\textbf{0.999}&\textbf{0.999}&\textbf{1.000}&\textbf{1.000}
\\
Keci-\approach 
&0.554&0.443&0.617&0.767
&0.919&0.880&0.952&0.974
\\
\bottomrule
\end{tabular}
\end{table*}

After these standard link prediction experiments, we delved into details to quantify the rate of overfitting (if any).
\Cref{table:train_fb15k237_yago310} reports the link prediction performance on the training splits of the two largest benchmark datasets.

These results indicate that
(1) SWA finds more accurate solutions than the conventional training 
(e.g. finding model parameters with Adam optimizer) 
and \approach
and (2) \textbf{maintaining a running average of model parameters can hinder generalization, as an underlying running model begins to overfit}.
Pertaining to (1), on YAGO3-10, SWA finds model parameters leading to higher link prediction performance on the training data in all metrics. 
Given that \ac{KGE} models have the same number of parameters (entities and relations are represented with $d$ number of real numbers), this superior performance of SWA over the conventional training can be explained as the mitigation of the noisy parameter updates around minima.
More specifically,
maintaining a running unweighted average of parameters (see \Cref{eq:reducing_lr_in_N_epoch}) particularly becomes particularly useful around a minima by means of reducing the noise in the gradients of loss w.r.t. parameters that is caused by the mini-batch training.

Pertaining to (2), \approach renders itself as an effective combination of SWA with early stopping techniques, where the former accepts all parameter updates on a parameter ensemble model based on a running model and the former rejects parameter updates on a running model in the presence of overfitting.
Since the \approach does not update a parameter ensemble model if a running model begins to overfit, it acts as a regularization on a parameter ensemble.
This regularization impact leads to a better generalization in all metrics.
\section{Discussion}

Our experimental results corroborate the findings of \cite{izmailov2018averaging}: Constructing a parameter ensemble model by maintaining a running average of parameters at each epoch interval improves the generalization performance across a wide range of datasets and models.
Link prediction, multi-hop query answering, and image classification results show that SWA and \approach find better solutions than conventional training based on ADAM and SGD optimizers. Our results also show that
updating the parameter ensemble uniformly on each epoch leads to sub-optimal results as an underlying model begins to overfit.
\approach effectively rejects parameter updates if an underlying model begins to overfit.
\approach renders itself as an effective combination of SWA with early stopping, where the former accepts all updates on a parameter ensemble model, while the latter rejects updates on a running model that begins to overfit.
Important, \approach does not require an initial starting epoch to start constructing a parameter ensemble.
Instead, the \approach performs a hard update on a parameter ensemble model if a running model outperforms a current parameter ensemble model on a validation dataset.
Yet, we observe that as the size of the validation dataset grows, the runtimes of \approach also grows, while the runtime performance of SWA is not influenced by the size of the validation dataset.
%
\section{Conclusion}

In this work, we investigated techniques to construct a high performing ensemble model, while alleviating the overhead of training multiple models, and retaining efficient memory and inference requirements at test time.
To this end, we propose an Adaptive Stochastic Weight Averaging (ASWA) technique that effectively combines the Stochastic Weight Averaging (SWA) technique with early stopping.
\approach extends SWA by building a parameter ensemble according to an adaptive schema governed by the generalization trajectory on the validation dataset.
\approach constructs a parameter ensemble model via its soft, hard, and reject updates.
Our extensive experiments over 11 benchmark datasets ranging from image classification to multi-hop query answering with 7 baselines indicate that \approach consistently improves generalization performances of baselines.
\approach more effectively alleviates overfitting than SWA across tasks.
\bibliography{example_paper}

\begin{thebibliography}{34}
\providecommand{\natexlab}[1]{#1}
\providecommand{\url}[1]{\texttt{#1}}
\expandafter\ifx\csname urlstyle\endcsname\relax
  \providecommand{\doi}[1]{doi: #1}\else
  \providecommand{\doi}{doi: \begingroup \urlstyle{rm}\Url}\fi

\bibitem[Arakelyan et~al.(2021)Arakelyan, Daza, Minervini, and Cochez]{arakelyan2021complex}
Erik Arakelyan, Daniel Daza, Pasquale Minervini, and Michael Cochez.
\newblock Complex query answering with neural link predictors.
\newblock In \emph{9th International Conference on Learning Representations, {ICLR} 2021, Virtual Event, Austria, May 3-7, 2021}. OpenReview.net, 2021.
\newblock URL \url{https://openreview.net/forum?id=Mos9F9kDwkz}.

\bibitem[Bala{\v{z}}evi{\'c} et~al.(2019)Bala{\v{z}}evi{\'c}, Allen, and Hospedales]{balavzevic2019tucker}
Ivana Bala{\v{z}}evi{\'c}, Carl Allen, and Timothy~M Hospedales.
\newblock Tucker: Tensor factorization for knowledge graph completion.
\newblock \emph{arXiv preprint arXiv:1901.09590}, 2019.

\bibitem[Baldi and Sadowski(2013)]{baldi2013understanding}
Pierre Baldi and Peter~J Sadowski.
\newblock Understanding dropout.
\newblock \emph{Advances in neural information processing systems}, 26, 2013.

\bibitem[Bishop and Nasrabadi(2006)]{bishop2006pattern}
Christopher~M Bishop and Nasser~M Nasrabadi.
\newblock \emph{Pattern recognition and machine learning}, volume~4.
\newblock Springer, 2006.

\bibitem[Breiman(1996)]{breiman1996bagging}
Leo Breiman.
\newblock Bagging predictors.
\newblock \emph{Machine learning}, 24:\penalty0 123--140, 1996.

\bibitem[Demir and Ngonga~Ngomo(2023)]{demir2023clifford}
Caglar Demir and Axel-Cyrille Ngonga~Ngomo.
\newblock Clifford embeddings--a generalized approach for embedding in normed algebras.
\newblock In \emph{Joint European Conference on Machine Learning and Knowledge Discovery in Databases}, pages 567--582. Springer, 2023.

\bibitem[Demir et~al.(2021)Demir, Moussallem, Heindorf, and Ngonga~Ngomo]{demir2021hyperconvolutional}
Caglar Demir, Diego Moussallem, Stefan Heindorf, and Axel-Cyrille Ngonga~Ngomo.
\newblock Convolutional hypercomplex embeddings for link prediction.
\newblock In Vineeth~N. Balasubramanian and Ivor Tsang, editors, \emph{Proceedings of The 13th Asian Conference on Machine Learning}, volume 157 of \emph{Proceedings of Machine Learning Research}, pages 656--671. PMLR, 17--19 Nov 2021.
\newblock URL \url{https://proceedings.mlr.press/v157/demir21a.html}.

\bibitem[Demir et~al.(2023)Demir, Wiebesiek, Lu, Ngonga~Ngomo, and Heindorf]{demir2023litcqd}
Caglar Demir, Michel Wiebesiek, Renzhong Lu, Axel-Cyrille Ngonga~Ngomo, and Stefan Heindorf.
\newblock Litcqd: Multi-hop reasoning in incomplete knowledge graphs with numeric literals.
\newblock In \emph{Joint European Conference on Machine Learning and Knowledge Discovery in Databases}, pages 617--633. Springer, 2023.

\bibitem[Dettmers et~al.(2018)Dettmers, Minervini, Stenetorp, and Riedel]{dettmers2018convolutional}
Tim Dettmers, Pasquale Minervini, Pontus Stenetorp, and Sebastian Riedel.
\newblock Convolutional 2d knowledge graph embeddings.
\newblock In \emph{Proceedings of the AAAI Conference on Artificial Intelligence}, volume~32, 2018.

\bibitem[Dietterich(2000)]{dietterich2000ensemble}
Thomas~G Dietterich.
\newblock Ensemble methods in machine learning.
\newblock In \emph{International workshop on multiple classifier systems}, pages 1--15. Springer, 2000.

\bibitem[Draxler et~al.(2018)Draxler, Veschgini, Salmhofer, and Hamprecht]{draxler2018essentially}
Felix Draxler, Kambis Veschgini, Manfred Salmhofer, and Fred Hamprecht.
\newblock Essentially no barriers in neural network energy landscape.
\newblock In \emph{International conference on machine learning}, pages 1309--1318. PMLR, 2018.

\bibitem[Gal and Ghahramani(2016)]{gal2016dropout}
Yarin Gal and Zoubin Ghahramani.
\newblock Dropout as a bayesian approximation: Representing model uncertainty in deep learning.
\newblock In \emph{international conference on machine learning}, pages 1050--1059. PMLR, 2016.

\bibitem[Garipov et~al.(2018)Garipov, Izmailov, Podoprikhin, Vetrov, and Wilson]{garipov2018loss}
Timur Garipov, Pavel Izmailov, Dmitrii Podoprikhin, Dmitry~P Vetrov, and Andrew~G Wilson.
\newblock Loss surfaces, mode connectivity, and fast ensembling of dnns.
\newblock \emph{Advances in neural information processing systems}, 31, 2018.

\bibitem[Goodfellow et~al.(2016)Goodfellow, Bengio, and Courville]{goodfellow2016deep}
Ian Goodfellow, Yoshua Bengio, and Aaron Courville.
\newblock \emph{Deep learning}.
\newblock MIT press, 2016.

\bibitem[He et~al.(2016)He, Zhang, Ren, and Sun]{he2016deep}
Kaiming He, Xiangyu Zhang, Shaoqing Ren, and Jian Sun.
\newblock Deep residual learning for image recognition.
\newblock In \emph{Proceedings of the IEEE conference on computer vision and pattern recognition}, pages 770--778, 2016.

\bibitem[Hinton et~al.(2012)Hinton, Srivastava, Krizhevsky, Sutskever, and Salakhutdinov]{hinton2012improving}
Geoffrey~E Hinton, Nitish Srivastava, Alex Krizhevsky, Ilya Sutskever, and Ruslan~R Salakhutdinov.
\newblock Improving neural networks by preventing co-adaptation of feature detectors.
\newblock \emph{arXiv preprint arXiv:1207.0580}, 2012.

\bibitem[Izmailov et~al.(2018)Izmailov, Podoprikhin, Garipov, Vetrov, and Wilson]{izmailov2018averaging}
Pavel Izmailov, Dmitrii Podoprikhin, Timur Garipov, Dmitry Vetrov, and Andrew~Gordon Wilson.
\newblock Averaging weights leads to wider optima and better generalization.
\newblock \emph{arXiv preprint arXiv:1803.05407}, 2018.

\bibitem[Krizhevsky et~al.(2009)Krizhevsky, Hinton, et~al.]{krizhevsky2009learning}
Alex Krizhevsky, Geoffrey Hinton, et~al.
\newblock Learning multiple layers of features from tiny images.
\newblock 2009.

\bibitem[Liu et~al.(2022)Liu, Chen, Atashgahi, Chen, Sokar, Mocanu, Pechenizkiy, Wang, and Mocanu]{liu2022deep}
Shiwei Liu, Tianlong Chen, Zahra Atashgahi, Xiaohan Chen, Ghada Sokar, Elena Mocanu, Mykola Pechenizkiy, Zhangyang Wang, and Decebal~Constantin Mocanu.
\newblock Deep ensembling with no overhead for either training or testing: The all-round blessings of dynamic sparsity.
\newblock In \emph{International Conference on Learning Representations}, 2022.
\newblock URL \url{https://openreview.net/forum?id=RLtqs6pzj1-}.

\bibitem[M{\"u}ller(2012)]{muller2012regularization}
Klaus-Robert M{\"u}ller.
\newblock Regularization techniques to improve generalization.
\newblock \emph{Neural Networks: Tricks of the Trade: Second Edition}, pages 49--51, 2012.

\bibitem[Murphy(2012)]{murphy2012machine}
Kevin~P Murphy.
\newblock \emph{Machine learning: a probabilistic perspective}.
\newblock MIT press, 2012.

\bibitem[Nickel et~al.(2011)Nickel, Tresp, and Kriegel]{nickel2011three}
Maximilian Nickel, Volker Tresp, and Hans-Peter Kriegel.
\newblock A three-way model for collective learning on multi-relational data.
\newblock In \emph{Icml}, 2011.

\bibitem[Polyak and Juditsky(1992)]{polyak1992acceleration}
Boris~T Polyak and Anatoli~B Juditsky.
\newblock Acceleration of stochastic approximation by averaging.
\newblock \emph{SIAM journal on control and optimization}, 30\penalty0 (4):\penalty0 838--855, 1992.

\bibitem[Prechelt(2002)]{prechelt2002early}
Lutz Prechelt.
\newblock Early stopping-but when?
\newblock In \emph{Neural Networks: Tricks of the trade}, pages 55--69. Springer, 2002.

\bibitem[Ren et~al.(2023)Ren, Galkin, Cochez, Zhu, and Leskovec]{ren2023neural}
Hongyu Ren, Mikhail Galkin, Michael Cochez, Zhaocheng Zhu, and Jure Leskovec.
\newblock Neural graph reasoning: Complex logical query answering meets graph databases.
\newblock \emph{arXiv preprint arXiv:2303.14617}, 2023.

\bibitem[Ruffinelli et~al.(2020)Ruffinelli, Broscheit, and Gemulla]{ruffinelli2020you}
Daniel Ruffinelli, Samuel Broscheit, and Rainer Gemulla.
\newblock You can teach an old dog new tricks! on training knowledge graph embeddings.
\newblock \emph{International Conference on Learning Representations}, 2020.
\newblock URL \url{https://openreview.net/forum?id=BkxSmlBFvr}.

\bibitem[Sagi and Rokach(2018)]{sagi2018ensemble}
Omer Sagi and Lior Rokach.
\newblock Ensemble learning: A survey.
\newblock \emph{Wiley Interdisciplinary Reviews: Data Mining and Knowledge Discovery}, 8\penalty0 (4):\penalty0 e1249, 2018.

\bibitem[Simonyan and Zisserman(2014)]{simonyan2014very}
Karen Simonyan and Andrew Zisserman.
\newblock Very deep convolutional networks for large-scale image recognition.
\newblock \emph{arXiv preprint arXiv:1409.1556}, 2014.

\bibitem[Srivastava et~al.(2014)Srivastava, Hinton, Krizhevsky, Sutskever, and Salakhutdinov]{srivastava2014dropout}
Nitish Srivastava, Geoffrey Hinton, Alex Krizhevsky, Ilya Sutskever, and Ruslan Salakhutdinov.
\newblock Dropout: a simple way to prevent neural networks from overfitting.
\newblock \emph{The journal of machine learning research}, 15\penalty0 (1):\penalty0 1929--1958, 2014.

\bibitem[Trouillon et~al.(2016)Trouillon, Welbl, Riedel, Gaussier, and Bouchard]{trouillon2016complex}
Th{\'e}o Trouillon, Johannes Welbl, Sebastian Riedel, {\'E}ric Gaussier, and Guillaume Bouchard.
\newblock Complex embeddings for simple link prediction.
\newblock In \emph{International conference on machine learning}, pages 2071--2080. PMLR, 2016.

\bibitem[Warde-Farley et~al.(2013)Warde-Farley, Goodfellow, Courville, and Bengio]{warde2013empirical}
David Warde-Farley, Ian~J Goodfellow, Aaron Courville, and Yoshua Bengio.
\newblock An empirical analysis of dropout in piecewise linear networks.
\newblock \emph{arXiv preprint arXiv:1312.6197}, 2013.

\bibitem[Xie et~al.(2013)Xie, Xu, and Chuang]{xie2013horizontal}
Jingjing Xie, Bing Xu, and Zhang Chuang.
\newblock Horizontal and vertical ensemble with deep representation for classification.
\newblock \emph{arXiv preprint arXiv:1306.2759}, 2013.

\bibitem[Xu et~al.(2021)Xu, Nayyeri, Vahdati, and Lehmann]{xu2021multiple}
Chengjin Xu, Mojtaba Nayyeri, Sahar Vahdati, and Jens Lehmann.
\newblock Multiple run ensemble learning with low-dimensional knowledge graph embeddings.
\newblock In \emph{2021 International Joint Conference on Neural Networks (IJCNN)}, pages 1--8. IEEE, 2021.

\bibitem[Yang et~al.(2015)Yang, Yih, He, Gao, and Deng]{yang2015embedding}
Bishan Yang, Wen-tau Yih, Xiaodong He, Jianfeng Gao, and Li~Deng.
\newblock Embedding entities and relations for learning and inference in knowledge bases.
\newblock In \emph{ICLR}, 2015.

\end{thebibliography}

\end{document}